\documentclass[]{mc_nankai}

\usepackage[dvipsnames, table, xcdraw]{xcolor}
\usepackage{utfsym}
\usepackage{booktabs}
\usepackage{amsmath, amssymb, mathtools, bm}
\usepackage{multirow, enumitem, subcaption, float, wrapfig}
\usepackage{pifont}
\usepackage{overpic}

\newcommand{\nMethod}{AgeBooth}

\title{\nMethod{}: Controllable Facial Aging and Rejuvenation via Diffusion Models}

\author[1]{Shihao Zhu$^{*}$}
\author[1]{Bohan Cao$^{*}$}
\author[1]{Ziheng Ouyang}
\author[2]{Zhen Li}
\author[3]{Peng-Tao Jiang}
\author[1]{Qibin Hou$^{\dagger}$}

\affiliation[1]{VCIP, School of Computer Science, Nankai University}
\affiliation[2]{The Chinese University of Hong Kong}
\affiliation[3]{vivo Mobile Communication Co., Ltd}

\affiliation[]{$^{*}$Equal Contribution.  $^{\dagger}$Corresponding author.}

\abstract{
Recent diffusion model research focuses on generating identity-consistent images from a reference photo, but they struggle to accurately control age while preserving identity, and fine-tuning such models often requires costly paired images across ages.
In this paper, we propose AgeBooth, a novel age-specific finetuning approach that can effectively enhance the age control capability of adapter-based identity personalization models without the need for expensive age-varied datasets.
To reduce dependence on a large amount of age-labeled data, we exploit the linear nature of aging by introducing age-conditioned prompt blending and an age-specific LoRA fusion strategy that leverages SVDMix, a matrix fusion technique. These techniques enable high-quality generation of intermediate-age portraits.
Our AgeBooth produces realistic and identity-consistent face images across different ages from a single reference image.
Experiments show that AgeBooth achieves superior age control and visual quality compared to previous state-of-the-art editing-based methods.
}

\mclink[Project Page]{\url{https://github.com/HH-LG/AgeBooth}}

\begin{document}

\maketitle
\justifying

\section{Introduction}\label{sec:intro}

Age transformation~\cite{guo_age_2024} is a computer vision task that modifies the apparent age of a person in a given facial image while preserving individual identity.
Based on the nature of the underlying methods, we categorize this task into two sub-tasks: face age editing and age controllable generation.
Face age editing refers to localizing facial modifications on existing images to reflect aging or rejuvenation while leaving all other aspects unchanged.
Most prior works~\cite{antipov2017face,yang2018age,song2018dualcgan,liu2019attribute,huang2020pfagan,farkhod2021reaging,liu2021a3gan,alaluf2021only,chen2023face}, especially those based on Generative Adversarial Networks (GANs)~\cite{goodfellow2014gan}, fall under this category.
Although they provide fine-grained age control, these methods face limitations, including training instability, visual artifacts, and limited flexibility in handling rich linguistic prompts.

The other sub-task is called age controllable generation.
It can also be seen as a sub-task of customized text-to-image (T2I) generation~\cite{gal2023textual,ruiz2023dreambooth,hu2022lora,kumari2023customdiffusion,wei2023elite,ye2023ipadapter}.
This task aims to generate a person’s image at a target age with fine-grained age control and identity preservation, while allowing contextual details to be created from textual prompts rather than retained from the original image.
Our proposed method falls into this category.
To achieve age controllable generation, we leverage the power of diffusion models~\cite{ho2020ddpm}, which have recently demonstrated remarkable capabilities in stable high-fidelity image synthesis and flexible conditioning.

We follow the work of tuning-free identity (ID) customization~\cite{ye2023ipadapter} to achieve ID consistency between generated images and reference images in our model.
ID customization models~\cite{ye2023ipadapter,li2024photomaker,xiao2024fastcomposer,guo2024pulid,jiang2025infiniteyou} achieve fine control over ID features to diffusion models by training adapters or finetuning but they often struggle to balance identity preservation and accurate age manipulation, as stronger identity preservation tends to stronger age preservation, thereby limiting age transformation. 
%

On the other hand, the limited image quality and insufficient diversity in age and identity in existing datasets hinder the finetuning of ID customization methods for age controllable generation.
For instance, AgeDB~\cite{moschoglou2017agedb} provides identity and age label with large time span but contains low-resolution images.
IMDB-Wiki~\cite{Rothe-ICCVW-2015} offers a large number of samples, but the majority are concentrated in similar age ranges.
FFHQ~\cite{karras2019style} contains high-resolution, diverse facial images but does not include identity labels.
To enable fine-tuning of ID customization models for generating images at specified ages, high-quality image pairs of the same identity at different ages are required, which is not feasible with the existing age datasets mentioned above.

Inspired by DreamBooth~\cite{ruiz2023dreambooth}, our proposed method is designed to operate in a few-shot setting to address these challenges, requiring only a small number of training images to achieve age effective controllable generation.
We introduce an age interpolation mechanism to support smooth transitions across different age stages.
We implement age transformation and identity preservation as independent modules, allowing the integration of various identity preservation models to achieve age control.

Specifically, the main contributions are summarized as follows:
\begin{itemize}
    \item We propose a finetuning paradigm that achieves effective age control even with limited training samples by leveraging lightweight tuning to bypass extensive data collection efforts.

    \item We introduce a dynamic fusion strategy for the age adapters and prompt tokens, enabling generation of unseen age groups without further finetuning, thus reducing the reliance on age-specific data.

    \item We demonstrate that our method, AgeBooth, can generate high-quality portraits across different age stages from a single reference image without requiring cross-age paired data, offering a plug-and-play extension that can integrate seamlessly with pre-trained ID-personalized generation models.
\end{itemize}

\begin{figure}[t!]
    \centering
    \includegraphics[width=0.9\textwidth]{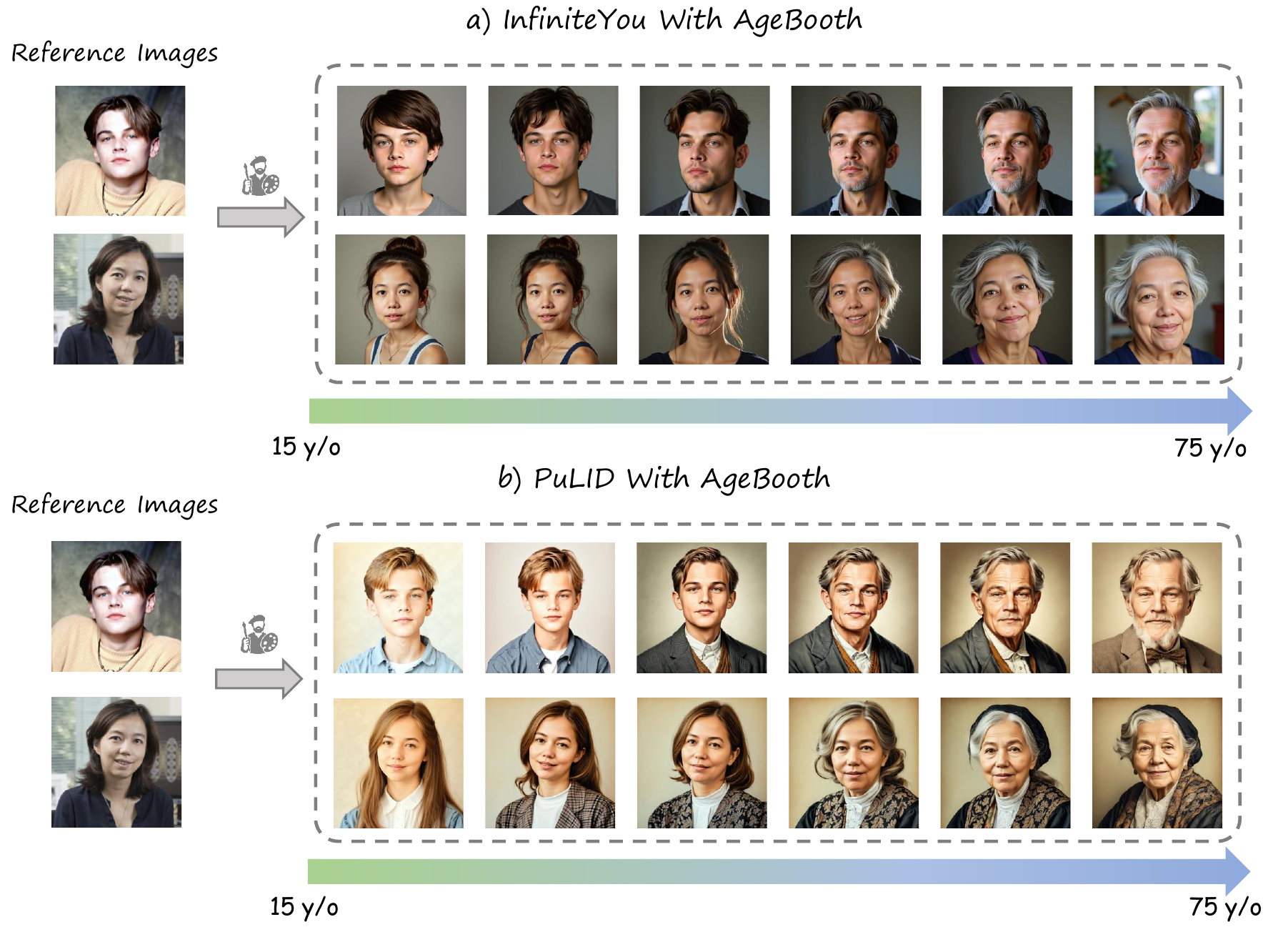}
    \caption{
    We propose an age transformation method, \textit{AgeBooth}, which can be applied to various Adapter-based ID customization models, enabling them to generate portraits of a person across different age stages: (a) shows a series of portraits generated by InfiniteYou with AgeBooth; (b) shows results generated by PuLID with AgeBooth using the same input.
    }
    \label{fig:teaser}
\end{figure}

\section{Related Work}\label{sec:related}

\textbf{Age Editing.}
Age editing is a process of digitally modifying a person's face in an image to appear older or younger while leaving all other visual features (e.g., identity, background, style, and posture) unchanged.
Early works, which form the majority of age editing research, are mostly based on GANs~\cite{antipov2017face,yang2018age,song2018dualcgan,liu2019attribute,huang2020pfagan,farkhod2021reaging,liu2021a3gan,alaluf2021only}.
These methods treat age as a conditional input and model the editing task as an image-to-image translation problem.
Most approaches achieve age editing by designing specialized generators and discriminators, such as pyramid architectures~\cite{yang2018age}, while some also incorporate attention mechanisms for more precise control~\cite{liu2021a3gan}.

With the emergence of latent diffusion models, several age editing methods have been proposed~\cite{chen2023face,li2023pluralistic,banerjee2024identity,hou2024high}, though they still face limitations in flexible text-based control. Contrast to these methods, our method achieves flexible text-based control by leveraging the original ID customization model.

\textbf{ID Customization.}
Identity customization models enable pre-trained models to generate images of user-specified identities guided by textual descriptions. 
Early approaches~\cite{hu2022lora,ruiz2023dreambooth,gal2023textual,kumari2023customdiffusion,zhou2024storydiffusion}, such as DreamBooth~\cite{ruiz2023dreambooth} and Textual Inversion~\cite{gal2023textual}, rely on finetuning the model for each input. 
Since this is computationally expensive during inference, tuning-free methods have gained popularity, including~\cite{guo2024pulid,wang2024instantid,ye2023ipadapter,li2024photomaker,xiao2024fastcomposer,he2024imagine,jiang2025infiniteyou,tao2025instantcharacter,paraperas2024arc2face}.
These methods typically use adapters, like CLIP~\cite{radford2021clip}, to inject identity information via cross-attention layers, significantly improving personalization efficiency.
Despite progress in ID customization, such methods often suffer from a copy-paste effect~\cite{guo2024pulid,jiang2025infiniteyou}, directly replicating the input face, which hinders performance in age transformation tasks.
Methods, like PhotoMaker~\cite{li2024photomaker}, generate results that are heavily influenced by the age depicted in the reference image, while in ControlNet-based approaches~\cite{wang2024instantid, zhang2024flashface} overly rigid ID control leads to even poorer age control.

Our method combines tuning-based and tuning-free strategies: Tuning teaches the model age progression, while adapters inject identity information to preserve facial characteristics. 
By decoupling age and identity modeling, our framework avoids copy-paste artifacts and enables smoother and more controllable age transformations.

\textbf{LoRA Fusion.}
LoRA fusion is a newer direction that focuses on combining two LoRA modules with different functions, aiming to preserve both capabilities during generation.
Some works~\cite{wu2024mixture,shah2024ziplora,frenkel2024implicit,zhang2023diffmorpher,ouyang2025k} explore merging LoRA layers using either learned (training-based) or heuristic (training-free) methods.
ZipLoRA~\cite{shah2024ziplora} learns mixing coefficients to merge two LoRA modules—one for style, one for content—into a single compressed LoRA.
DiffMorpher~\cite{zhang2023diffmorpher} creates videos with smooth transitions by interpolating LoRA weights and attention maps.
Different from these methods, which focus on style-content fusion or smooth transitions, our approach uses LoRA fusion to support age-controllable image generation across a wide age range, ensuring both age consistency and image quality.

\section{Method}\label{sec:methods}

\begin{figure*}[t!]
    \centering
    \includegraphics[width=0.9\textwidth]{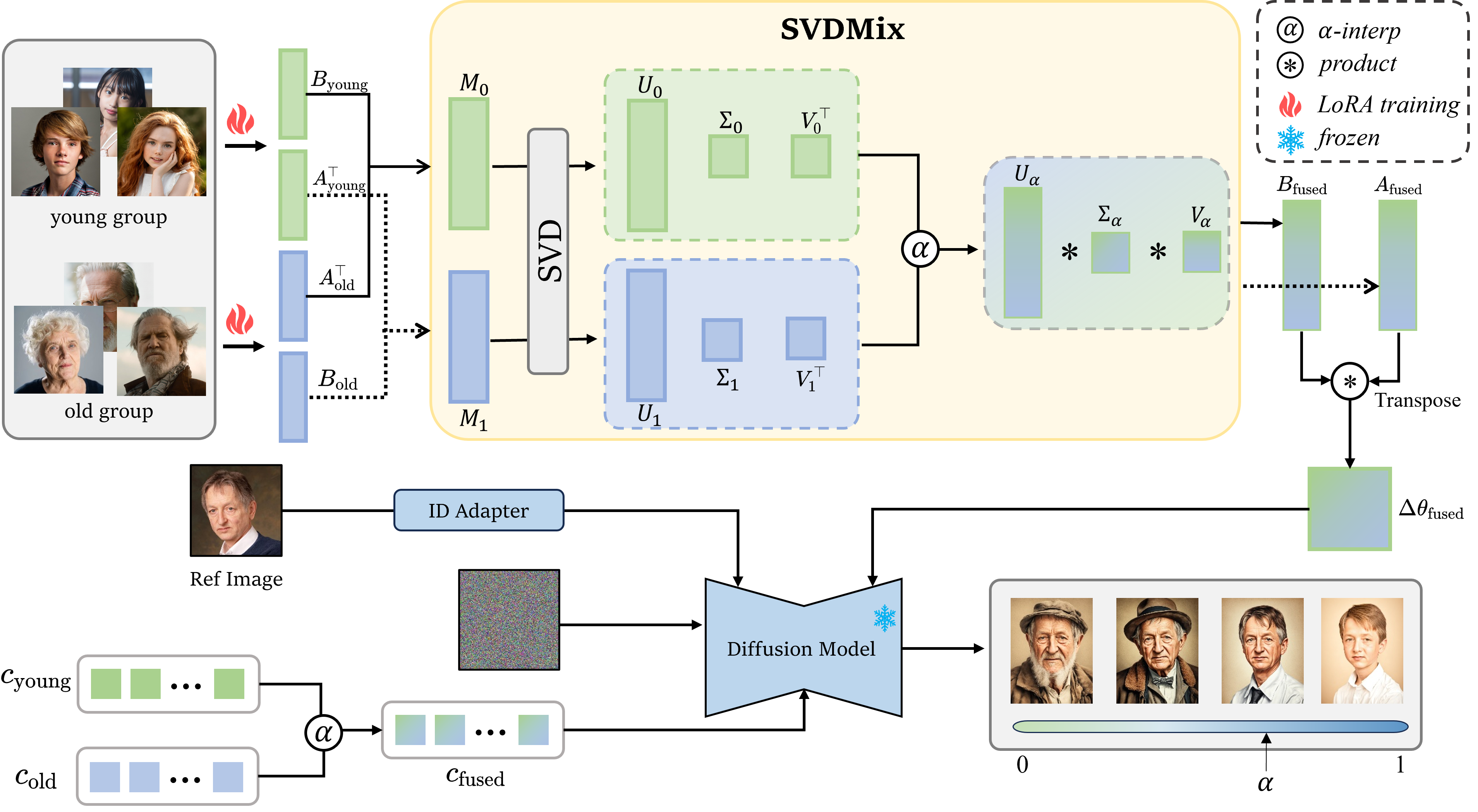}
    \caption{
    \textbf{Overview of the AgeBooth framework}:  
    AgeBooth first performs LoRA fine-tuning on both young and old age groups, enabling the model to learn the concepts of youth and aging. 
    The resulting LoRA modules for young and old are then fused using the proposed SVD-based method and applied to the base model.
    To preserve identity and further control the generated age via prompts, a pre-trained ID Adapter is integrated into the model, and the input prompts are interpolated accordingly.
    }
    \label{fig:method}
\end{figure*}

In this section, we first introduce some preliminaries about ID adapters for diffusion and Low-Rank Adaptation (LoRA), which serve as the foundation of our method.
We then elaborate on our proposed framework, which consists of two sequential components, as shown in Fig.\ref{fig:method}.  
The first component, Few-Shot Age-Specific Finetuning, enables the model to learn age concepts from a few examples while preserving subject identity via an integrated ID adapter.
The second component, Training-Free LoRA \& Prompt Fusion, allows for smooth cross-age generation by interpolating age-specific LoRA weights and prompts without any additional training.

\subsection{Preliminaries}
\textbf{ID adapters for diffusion}~\cite{guo2024pulid,ye2023ipadapter,wang2024instantid,jiang2025infiniteyou} are designed to preserve and manipulate subject-specific identity information during image generation.
When integrated with pretrained diffusion models~\cite{ho2020ddpm}, ID adapters enable personalized image synthesis by conditioning the generation process on identity embeddings.



To incorporate identity information into each cross-attention~\cite{vaswani2017attention} layer $l$, ID adapters inject identity embeddings $\boldsymbol{c_I}$ via a dedicated cross-attention mechanism operating on the noisy latent representation $\boldsymbol{x}_t$.
At each layer $ l $, two separate attention maps are computed using the standard scaled dot-product attention~\cite{vaswani2017attention}:
\begin{equation}
\operatorname{Attn}(\boldsymbol{q}, \boldsymbol{k}, \boldsymbol{v}) = \operatorname{softmax}\left( \frac{\boldsymbol{q} \boldsymbol{k}^\top}{\sqrt{d}} \right) \boldsymbol{v},
\end{equation}
where the query $\boldsymbol{q}$ is obtained by applying a linear projection to the current latent $\boldsymbol{x}_t$, i.e., $\boldsymbol{q} = \boldsymbol{W}^{(l)}_q \boldsymbol{x}_t$, and the keys/values are projected from the text embedding $\boldsymbol{c_T}$ and the identity embedding $\boldsymbol{c_I}$ as
$\boldsymbol{k}_T = \boldsymbol{W}^{(l)}_k \boldsymbol{c_T}, \, \boldsymbol{v}_T = \boldsymbol{W}^{(l)}_v \boldsymbol{c_T}$ and
$\boldsymbol{k}_I = \boldsymbol{W}^{(l)}_{k_\text{id}} \boldsymbol{c_I}, \, \boldsymbol{v}_I = \boldsymbol{W}^{(l)}_{v_\text{id}} \boldsymbol{c_I}$.
The final conditioning context at layer $l$ is obtained by a weighted combination of the two attention outputs:
\begin{equation}\label{eq:id_modulation}
\operatorname{Attn}_{\text{fused}}^{(l)} = \operatorname{Attn}^{(l)}(\boldsymbol{q}, \boldsymbol{k}_T, \boldsymbol{v}_T) + \gamma \cdot \operatorname{Attn}^{(l)}(\boldsymbol{q}, \boldsymbol{k}_I, \boldsymbol{v}_I),
\end{equation}
where $\gamma$ is a hyperparameter that modulates the contribution of identity-specific information in the fused attention.

\textbf{LoRA} (Low-Rank Adaptation)~\cite{hu2022lora} was originally proposed as a technique for finetuning large-scale language models.
When applied to pretrained T2I diffusion models~\cite{podell2024sdxl,ramesh2022hierarchical,saharia2022photorealistic}, LoRA enables the generation of high-quality stylized images.
Its remarkable efficiency has made it a widely adopted approach in the finetuning of such models.
The central assumption behind LoRA is that the parameter update matrix $\Delta \theta \in \mathbb{R}^{m \times n}$, derived during the finetuning of large models, contains many negligible or near-zero elements. 
This allows $\Delta \theta$ to be approximated using a low-rank decomposition into two smaller matrices: $B \in \mathbb{R}^{m \times r}$ and $A \in \mathbb{R}^{r \times n}$, such that $\Delta \theta = BA$.
Here, $r$ is the intrinsic rank of $\Delta \theta$, satisfying $r \ll \min(m, n)$. 
During training, the original model weights $\theta_0 \in \mathbb{R}^{m \times n}$ remain frozen, while only $A$ and $B$ are updated. 
The resulting finetuned parameters are given by $\theta = \theta_0 + BA$.
This parameter-efficient strategy enables adaptation to downstream tasks by modifying only a small subset of the total parameters.
Due to this efficiency, LoRA has seen widespread use in finetuning T2I diffusion models.

\subsection{Few-Shot Age-Specific Finetuning}\label{sec:age-finetuning}

In this subsection, we present our method for guiding a model to generate age-specific images without relying on large, labeled datasets.  
Given a target age $x$, we finetune a base T2I model $\boldsymbol{\epsilon}_\theta$ using a small set of images of individuals at age $x$, enabling the model to internalize and reproduce the concept of that age during inference.

Formally, given a small dataset $ \mathcal{D}_x = \{(\boldsymbol{x}_0^{(i)}, \boldsymbol{p}_x)\}_{i=1}^N $ where each image $ \boldsymbol{x}_0^{(i)} $ depicts a person aged $ x $ (e.g., 10-20 years old), and $ \boldsymbol{p}_x $ is a prompt containing the unique identifier for this age, such as ``\texttt{A person in sbu age}'', we finetune a pretrained text-to-image model $ \boldsymbol{\epsilon}_\theta $ using the DreamBooth~\cite{ruiz2023dreambooth} framework.
Here, ``\texttt{sbu}'' is a commonly used placeholder in DreamBooth training, representing a specific feature to establish a correspondence between the prompt and the concept being learned.
Specifically, we optimize the model parameters $ \theta $ by minimizing the following loss function:
\begin{equation}
\mathcal{L}_{\text{DM}} = \mathbb{E}_{\boldsymbol{x}_0, \boldsymbol{\epsilon} \sim \mathcal{N}(\mathbf{0}, \mathbf{I}), \boldsymbol{c_T}, t}\left\|\boldsymbol{\epsilon}-\boldsymbol{\epsilon}_\theta\left(\boldsymbol{x}_t, \boldsymbol{c_T}, t\right)\right\|^2,
\end{equation}
where $\boldsymbol{x}_0$ is the original image sampled from the training set $\mathcal{D}_x$, and $\boldsymbol{x}_t$ is its noisy version at diffusion timestep $t$ in the forward process.
The conditioning vector $\boldsymbol{c_T}$ is obtained by encoding the prompt $\boldsymbol{p}_x$ using a text encoder $\Gamma$, i.e., $\boldsymbol{c_T} = \Gamma(\boldsymbol{p}_x)$. The loss encourages the model to perform accurate denoising while being conditioned on the learned age-specific concept. 

By training only on a small number of age-specific images ($N \approx 10$) and conditioning on ``\texttt{sbu age.}'', the model learns to associate this phrase with the visual appearance of the target age group as shown in Fig.~\ref{fig:method}. 
However, while the finetuned model captures age-related visual features, it does not inherently support identity conditioning and thus cannot preserve subject identity across generations.
To address this, we integrate a pretrained ID adapter into the T2I model, enabling it to accept identity embeddings as input.
As formulated in Eqn.~\ref{eq:id_modulation}, the ID adapter injects identity embeddings during generation, allowing the model to synthesize age-consistent images while maintaining the visual identity of the subject.

\subsection{Training-free LoRA and Prompt Fusion}\label{subsec:fusions}

Building upon the above finetuning approach, we can obtain a collection of LoRA adapters $\{\Delta\theta_x\}$, each corresponding to a specific age group $x$.
These adapters allow the model to synthesize individuals at various age stages by finetuning on a small set of age-specific images per concept.
However, covering a broad and continuous age spectrum would still require collecting and finetuning on multiple age groups, which becomes impractical under limited high-quality data.
To address this, we propose a training-free strategy that leverages both LoRA and prompt fusion to interpolate between age groups.
Specifically, we combine two representative age-specific LoRA adapters—$\Delta\theta_\text{young}$ and $\Delta\theta_\text{old}$—along with their associated prompts, enabling smooth and continuous age transitions without the need for additional training.

\textbf{Naive linear LoRA fusion.}
We first observe that a simple linear interpolation between $\Delta\theta_\text{young}$ and $\Delta\theta_\text{old}$ can already produce good images with gradual changes in perceived age.
This interpolation is defined as:
\begin{equation}
\Delta\theta_{\text{fused}} = \alpha \cdot \Delta\theta_\text{young} + (1 - \alpha) \cdot \Delta\theta_\text{old},
\end{equation}
where $\alpha \in [0, 1]$ controls the blending ratio and effectively the target age.
Despite its simplicity, this approach sometimes leads to suboptimal visual quality or attribute drift, especially when the two LoRA adapters encode significantly different features, as shown in Fig.~\ref{fig:ablations}.

\textbf{SVD-based LoRA fusion.}
To further enhance the fidelity and consistency of interpolation results, we propose a fusion strategy based on Singular Value Decomposition (SVD), inspired by~\cite{liu2025onepromptonestory,li2024get,gu2014weighted}.
This method leverages the low-rank structure of LoRA layers to achieve high-quality fusion while maintaining computational efficiency.

Given two matrices of the same shape $M_0, M_1 \in \mathbb{R}^{m \times n}$, we introduce a fusion operation based on singular value decomposition, which we denote as SVDMix, defined as:
\begin{equation}
\operatorname{SVDMix}(M_0, M_1; \alpha) = U_{\alpha} \Sigma_{\alpha} V_{\alpha}^\top,
\end{equation}
where each matrix $M_i$ is decomposed as:
\begin{equation}
M_i = U_i \Sigma_i V_i^\top, \quad i \in \{0, 1\},
\end{equation}
and the fused components are computed via linear interpolation:
\begin{equation}
X_{\alpha} = \alpha X_0 + (1 - \alpha) X_1, \quad X \in \{U, \Sigma, V\}.
\end{equation}

Assume that age-specific finetuning has been conducted, resulting in LoRA weight updates for different age groups:
\begin{equation}
\Delta\theta_* = B_* A_*, \quad * \in \{\text{young}, \text{old}\},
\end{equation}
where $B_* \in \mathbb{R}^{m \times r}$ and $A_* \in \mathbb{R}^{r \times n}$, with $r \ll \min(m, n)$. 
This follows the standard low-rank decomposition used in LoRA.

While it is possible to apply SVDMix directly to the full LoRA matrices $\Delta\theta_* \in \mathbb{R}^{m \times n}$, this incurs a computational complexity of $O(\min(m, n)^2 \cdot \max(m, n))$, which is expensive for modern deep diffusion models with large dimensions (e.g., $m = n = 1024$).
To address this, we instead apply SVDMix separately to the smaller projection matrices $B_*$ and $A_*$, which reduces the complexity to $O((m + n)r^2)$. 
This leads to a more efficient fusion procedure without sacrificing quality.
The fusion of the projection matrices can be performed as:
\begin{align}
B_{\text{fused}} &= \operatorname{SVDMix}(B_\text{young}, B_\text{old}; \alpha),
 \\
A_{\text{fused}} &= \operatorname{SVDMix}(A_\text{young}, A_\text{old}; \alpha).
\end{align}

The final fused LoRA weight is then reconstructed as:
\begin{equation}
\Delta\theta_{\text{fused}} = B_{\text{fused}} A_{\text{fused}}.
\end{equation}

This SVD-based fusion approach not only provides smooth and stable interpolation in the LoRA feature space, but also ensures computational efficiency and full compatibility with the standard LoRA design (as shown in the ablation study in Fig.~\ref{fig:ablations}).
A rigorous proof that the proposed fusion method defines a continuous mapping is provided in Appendix~\ref{sec:continuity}.

\textbf{Prompt fusion.}
In parallel with LoRA weight fusion, we adopt a similar method to~\cite{zhang2023diffmorpher} by interpolating between age-related textual prompts to further refine the generated output.
Given two prompts $\boldsymbol{p}_{\text{young}}$ and $\boldsymbol{p}_{\text{old}}$, we compute their embeddings using a text encoder $\Gamma$, denoted as $\boldsymbol{c}_\text{young} = \Gamma(\boldsymbol{p}_\text{young})$ and $\boldsymbol{c}_\text{old} = \Gamma(\boldsymbol{p}_\text{old})$. We then perform linear interpolation in the embedding space:
\begin{equation}
\boldsymbol{c}_\text{fused} = \alpha \cdot \boldsymbol{c}_\text{young} + (1 - \alpha) \cdot \boldsymbol{c}_\text{old},
\end{equation}
where $\alpha \in [0, 1]$ controls the interpolation ratio.
This fused prompt embedding is then used as the conditioning input during inference, aligning semantic guidance with the interpolated visual style.
By jointly fusing LoRA weights and prompts, our method enables continuous, high-quality, and identity-consistent age progression or regression without additional training or annotations.

\section{Experiments}\label{sec:exp}

\begin{figure}[!htbp]
    \centering
    \includegraphics[width=0.8\textwidth]{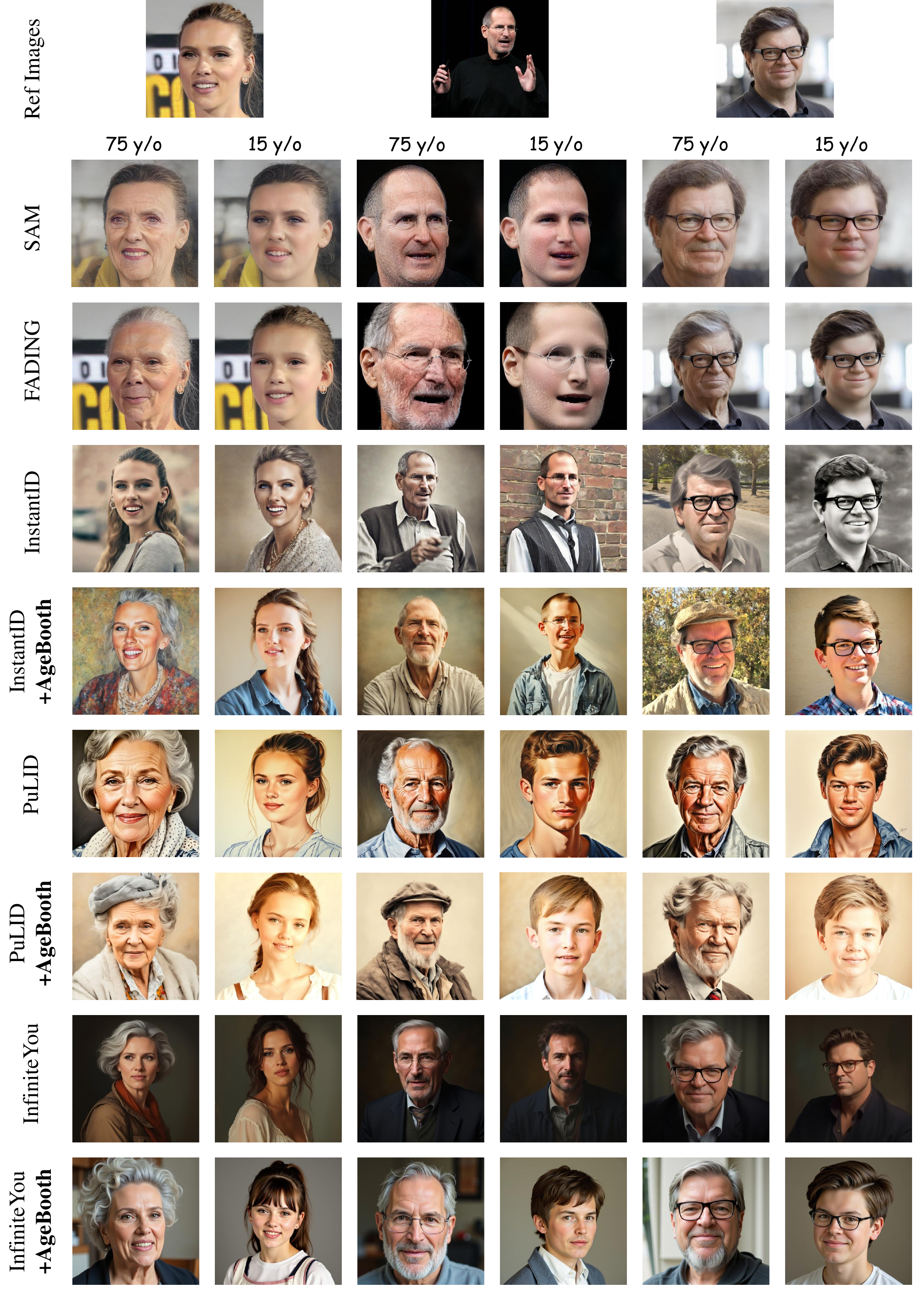}
    \caption{
    Comparisons of identity-personalized results with and without our proposed AgeBooth, alongside existing age editing methods.
    }
    \label{fig:qualitative_exp}
\end{figure}

\subsection{Implementation Details}\label{subsec:implementation}

\textbf{Dataset collection.}
To evaluate our method's performance in age-specific portrait style transfer, we construct a customized dataset focusing on two representative age groups: 10-20 and 70-80 years old.
We first filter samples from the IMDB-WIKI~\cite{Rothe-IJCV-2018,Rothe-ICCVW-2015} dataset based on age labels and then manually select around 20 high-quality face images per group, considering factors, such as facial quality, pose, and expression.
To increase diversity and realism, we further collect over 10 additional images per group from publicly available online sources, ensuring high resolution, frontal or near-frontal poses, minimal occlusion, and balanced identity and gender distribution.
Our choice of age \emph{groups} rather than exact ages is discussed in Appendix~\ref{app:agegroup}, where we quantify label noise and show that using perfectly aligned ages yields negligible gains.

\textbf{Experiment details.}
We implement and evaluate the proposed AgeBooth on several base models, including InstantID~\cite{wang2024instantid}, PuLID-v1.1~\cite{guo2024pulid}, and InfiniteYou-FLUX v1.0~\cite{jiang2025infiniteyou} (aes\_stage2 version with photorealistic LoRA enabled). 
Finetuning experiments are conducted respectively on SDXL~\cite{rombach2022high} and FLUX.1-dev~\cite{flux2024}, with detailed strategies described in Sec.~\ref{sec:age-finetuning}.
For SDXL finetuning, we use Juggernaut-XL-v9 as the base model. 
The LoRA rank is set to 16, and training images are at a resolution of 512$\times$512. 
We employ the AdamW optimizer with an initial learning rate of 1e-5, a batch size of 1, and a maximum of 1000 training steps.
The prompt used during training is consistently ``\texttt{A person in sbu age}''.
For FLUX.1-dev fine-tuning, the same prompts and training steps as SDXL are retained, but the optimizer is replaced with Prodigy, and the learning rate is set to 1.
Additionally, the rank of the LoRA is reduced to 4.
All the training processes are conducted on a single NVIDIA A40 GPU.

\subsection{Qualitative Comparison}

As illustrated in Fig.~\ref{fig:qualitative_exp}, AgeBooth is applied to InstantID~\cite{wang2024instantid}, PuLID~\cite{guo2024pulid}, and InfiniteYou~\cite{jiang2025infiniteyou}, and compared with the age editing methods SAM~\cite{alaluf2021matter} and FADING~\cite{chen2023face}.
The age control of the base ID customization models is achieved by inserting precise descriptions of target age and age-related human descriptors (e.g., boy, old man) into the text prompts.
As shown in the 1st and 2nd rows, age editing methods can generate relatively accurate ages, but often fail to adjust age-specific features such as facial shape and clothing, leading to identity loss (e.g., 1st, 4th, 5th, and 6th columns) and lower image quality. 
In contrast, the 3rd, 5th, and 7th rows demonstrate that identity-customized models produce higher-quality and more diverse results, but tend to preserve too much from the reference image, resulting in inaccurate age portrayal—such as elderly subjects without white hair or young individuals appearing overly mature.
Our method (4th, 6th, and 8th rows) effectively combines the strengths of both approaches, generating accurate age-specific features, changes in hair color, facial structure, and even age-appropriate clothing, while achieving higher image quality and greater diversity.
More examples generated by AgeBooth are provided in Appendix~\ref{app:visual_res}.

\begin{table}[t!]
\centering
\caption{ \textbf{Age accuracy comparisons.}
The evaluated metrics include mean age ($\overline{\text{Age}}$) and the mean absolute error (MAE) of the generated ages.
The absolute difference between the estimated mean age and the target age, denoted as $|\delta|$, is annotated at the bottom right of the corresponding mean age.
The numbers 15 and 75 represent two representative age ranges selected for evaluation.
The best results are shown in \textbf{bold}, while the second-best are \underline{underlined}.
}
\label{tab:quantity}
\small
\begin{tabular}{@{}ccccccc@{}}
\toprule
 & 15 $\overline{\text{Age}}_{|\delta|\downarrow}$ & 15 MAE$\downarrow$ &
  45 $\overline{\text{Age}}_{|\delta|\downarrow}$ & 45 MAE$\downarrow$ &
 75 $\overline{\text{Age}}_{|\delta|\downarrow}$ & 75 MAE$\downarrow$ \\ \midrule
SAM & $18.99_{3.99}$ & 4.61 & $ 47.23_{2.23}$ & 4.47 & $\underline{69.84}_{\underline{5.16}}$ & \underline{6.44} \\
FADING & $\textbf{14.20}_\textbf{0.80}$ & \textbf{2.44} & $\underline{43.02}_{\underline{1.98}}$ & 9.57 & $64.86_{10.14}$ & 12.57 \\
InstantID & $22.33_{7.33}$ & 7.35 &  $ 22.34_{22.66}$ & 22.66 &  $37.83_{37.17}$ & 37.17 \\
\begin{tabular}[c]{@{}c@{}}InstantID\textit{+AgeBooth}\end{tabular} & $19.61_{4.61}$ & 4.92 & $29.13_{15.87}$ & 16.9 & $50.90_{24.10}$ & 24.14 \\
PuLID & $24.13_{9.13}$ & 9.13 & $40.81_{4.19}$ & \underline{4.43} & $50.37_{24.63}$ & 24.66 \\
\begin{tabular}[c]{@{}c@{}}PuLID\textit{+AgeBooth}\end{tabular} & $\underline{16.79}_{\underline{1.79}}$ & \underline{2.82} &  $\textbf{43.74}_\textbf{1.26} $& \textbf{4.32} & $\textbf{78.82}_{\textbf{3.82}}$ & \textbf{3.98} \\
InfiniteYou & $26.38_{11.38}$ & 11.38 &   ${40.97}_{4.03}$& 16.52 & $44.22_{30.78}$ & 30.77 \\
\begin{tabular}[c]{@{}c@{}}InfiniteYou\textit{+AgeBooth}\end{tabular} & $18.50_{3.50}$ & 4.17 & $38.36_{6.64}$ & 12.11 & $51.82_{23.18}$ & 23.18 \\ 
\bottomrule
\end{tabular}

\end{table}

\begin{table}[!t]
\centering
\small
\caption{\centering \textbf{Aesthetic comparisons}.Scores of LAION predictor.}
\label{tab:aesthetic}
\begin{tabular}{@{}lccccc@{}}
\toprule
\textbf{Method} & SAM & FADING & InstantID+\textit{AgeBooth} & PuLID+\textit{AgeBooth} & InfiniteYou+\textit{AgeBooth} \\ \midrule
\textbf{Aesthetic Score} & 6.21 & 6.07 & \textbf{7.04} & \underline{7.03} & 6.53 \\
\bottomrule
\end{tabular}
\end{table}

\begin{table*}[t]
\centering
\small
\caption{\centering Image quality comparisons on different Interpolation methods as $\alpha$ changes.
}
\label{tab:interp_ablation}
\begin{tabular}{@{}ccccccccccc@{}}
\toprule
$\alpha$ &
  0.10 &
  0.20 &
  0.30 &
  0.40 &
  0.50 &
  0.60 &
  0.70 &
  0.80 &
  0.90 &
  Avg. \\ \midrule
naive linear &
  7.26 &
  7.33 &
  7.54 &
  7.11 &
  6.83 &
  6.60 &
  6.47 &
  6.36 &
  6.40 &
  6.88 \\
SVD based &
  \textbf{7.72} &
  \textbf{7.70} &
  \textbf{7.95} &
  \textbf{7.34} &
  \textbf{7.14} &
  \textbf{6.96} &
  \textbf{7.11} &
  \textbf{7.25} &
  \textbf{6.42} &
  \textbf{7.29} \\ \bottomrule
\end{tabular}
\end{table*}

\subsection{Quantitative Comparisons}

Following~\cite{chen2023face,gomez2022custom}, we evaluate the performance of our method using tasks similar to theirs: \textit{Input}~$\rightarrow$~\textit{X}, where \textit{X} corresponds to three representative target age groups: 15, 45, and 75 years.
We randomly select images from CelebHQ~\cite{karras2018progressive} as our test dataset.
We use the MiVOLO~\cite{mivolo2023,mivolo2024} age estimator to detect the age of the generated images and compute the average error between the estimated and target ages, which is reported in Tab.~\ref{tab:quantity}.
As shown in the table, our method significantly improves the age accuracy of various base models, achieving the best performance when applied to the PuLID base model.
In terms of overall age accuracy, our method surpasses the previous state-of-the-art age editing approach, FADING~\cite{chen2023face}.

To highlight the superior image quality of our method over age editing approaches, we computed average aesthetic scores using the LAION predictor~\cite{laionaes} in Tab.~\ref{tab:aesthetic}.
Our method achieves noticeably higher scores, which we attribute to its ability to generate results that are more natural and visually pleasing, thanks to being less constrained by the structural limitations of the original image.

\begin{figure}[!t]
    \centering
    \includegraphics[width=0.8\textwidth]{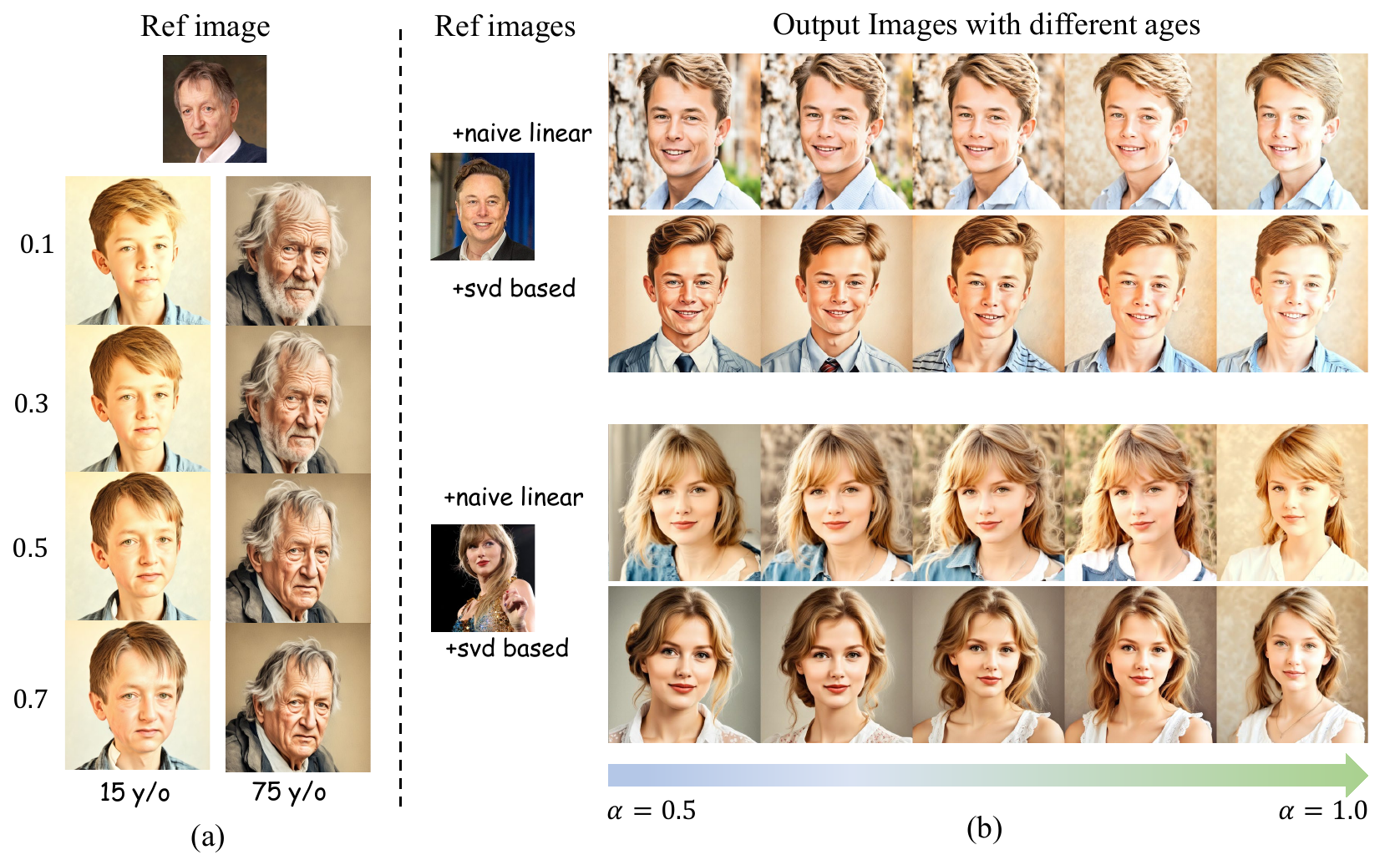}
    \caption{\textbf{Ablation study.}
    (a) Impact of different identity modulation factor $\gamma$ of the pretrained ID adapter.
    (b) Evaluations of different LoRA interpolation methods introduced in Sec.~\ref{subsec:fusions}.
    }
    \label{fig:ablations}
    
\end{figure}

\subsection{Ablation Study}

\textbf{Identity modulation factor $\gamma$.}
Our method leverages a pretrained ID adapter by injecting controllable identity features through the addition of the original attention and identity attention in Eq.~\ref{eq:id_modulation}.
By adjusting the identity modulation factor $\gamma$, we can balance the trade-off between identity consistency and age control.
To determine the optimal value of $\gamma$, we conducted experiments using the pretrained ID adapter provided by PuLID.
As shown in Fig.~\ref{fig:ablations}, we found that a $\gamma$ of 0.1 fails to provide sufficient identity information, while a $\gamma$ of 0.5 introduces excessive age-related details—such as wrinkles and beards—into the generated images.
In practice, we set $\gamma$ to 0.3 to achieve a balance between preserving identity and controlling age features, thus yielding the best performance in both aspects.

\textbf{Different LoRA interpolation methods.}
Fig.~\ref{fig:ablations} also shows the results of using naive linear interpolation and SVD-based interpolation in LoRA interpolation.
It can be observed that while naive linear interpolation allows the model to generate images of different ages as the $\alpha$ changes, it often leads to blurred and cluttered backgrounds, including artifacts such as wrinkles and other noise.
In contrast, the proposed SVD-based interpolation significantly improves image quality and yields better background consistency.
This qualitative improvement is corroborated by the quantitative results reported in Tab.~\ref{tab:interp_ablation}, where the SVD-based method consistently outperforms naive interpolation across most $\alpha$ values and achieves a higher overall average score.

\section{Conclusion}\label{sec:conclusion}
This paper presents AgeBooth, a novel few-shot approach that introduces age control into the ID customization task within text-to-image generation.
AgeBooth equips the model with age generation capability in a few-shot setting by using age-specific LoRA as well as LoRA \& Prompt Fusion, without compromising ID preservation.
Experiments show that AgeBooth surpasses existing approaches, enabling more flexible and precise age-controllable personalized image generation.
Future work could focus on improving age control accuracy and exploring its integration into broader personalized generation tasks.

\textbf{Limitations.}
Although our method AgeBooth significantly improves age controllability with only a few training images, it still has some limitations. 
Due to the interpolation operation, the age accuracy of images generated by the fused LoRA may degrade.
Despite this potential degradation, our method still achieves improvements over previous approaches.
To mitigate this issue, one potential solution is to fit a function that maps the interpolation coefficient $\alpha$ to the generated age, enabling more linear and precise control.
We will further address these issues in our future work.


\bibliography{egbib}
\bibliographystyle{ieee_fullname}

\clearpage
\newpage
\appendix

\section{Why Age Groups?}\label{app:agegroup}

\paragraph{(i) Tight filtering around the target age.}
For each target (e.g.\ 15\,yr), we re-estimate the biological age of every candidate image with MiVOLO~\cite{mivolo2023,mivolo2024} and keep only those whose error $\lvert\hat{Age} - Age_{\text{target}}\rvert \le 3$\,yr, the empirical RMSE of our pipeline.  
Thus the so-called ``10–20'' range in the main text is, in practice, centered tightly on 15.

\paragraph{(ii) Upper bound given by estimator noise.}
State-of-the-art age estimators, \emph{e.g.}, MiVOLO~\cite{mivolo2023,mivolo2024}, report a 3.65\,yr MAE on in-the-wild faces.  
Therefore, even “exact” labels still carry $\ge\!3$\,yr latent noise, limiting the value of pursuing single-year annotation.

\paragraph{(iii) Exact-age ablation.}
We repeated fine-tuning under three alternative label settings:
(a) the default \textbf{age group} filtering (±3 yr);
(b) a much smaller subset whose \textbf{ground truth} annotation is exactly 15 / 45 / 75 yr;
(c) a subset whose \textbf{estimated} age (via MiVOLO) falls exactly on those targets.

\begin{table}[!htbp]
\centering
\caption{Age errors ($|\delta|$, MAE) for age group, GT-exact, and estimator-exact training.}
\label{tab:why_group}
\begin{tabular}{@{}cccc@{}}
\toprule
                & age group & ground truth & estimated\\ \midrule
15 ${|\delta|}$ & \textbf{4.61}      & 10.44        & 7.61     \\
15 MAE          & \textbf{4.92}      & 10.90        & 8.28     \\
45 ${|\delta|}$ & 15.87     & 8.42         & \textbf{6.69}     \\
45 MAE          & 16.90     & 13.61        & \textbf{13.56}    \\
75 ${|\delta|}$ & 24.10     & \textbf{23.03}        & 26.06    \\
75 MAE          & \textbf{24.14}     & 24.86        & 26.58    \\ \bottomrule
\end{tabular}
\end{table}

Tab.~\ref{tab:why_group} indicates that refining to \emph{ground-truth} 15-yr faces offers virtually no accuracy gain over our age-group baseline, while the subset defined by \emph{estimated} exact ages yields a noticeable improvement only at the 45-yr target; for the other two targets the age-group setting remains superior. Given the negligible benefit at 15 yr, the single-age advantage limited to 45 yr, and the substantially higher curation cost of assembling exact-age subsets, we retain the age-group strategy as the most cost-effective choice.

These results confirm that age groups strike a better balance between annotation effort and performance.

\section{Continuity of the SVDMix}\label{sec:continuity}

This section establishes that the fused LoRA weight $\Delta\theta_{\text{fused}}(\alpha)\in\mathbb R^{m\times n}$ obtained with the SVD\,-\,based fusion strategy is a \emph{continuous} function of the blending coefficient $\alpha\in[0,1]$.


Let $B_{\text{young}},B_{\text{old}}\in\mathbb R^{m\times r}$ and $A_{\text{young}},A_{\text{old}}\in\mathbb R^{r\times n}$ with $r\ll\min\{m,n\}$.
For each $\ast\in\{\text{young},\text{old}\}$ we write the thin singular value decomposition (SVD) as  
\begin{equation}
  B_{\ast}=U_{\ast}\Sigma_{\ast}V_{\ast}^{\!\top},
  \quad
  A_{\ast}=P_{\ast}\Lambda_{\ast}Q_{\ast}^{\!\top},
  \label{eq:svd}
\end{equation}
where the factors have the usual orthogonality and diagonality properties.
For a given $\alpha\in[0,1]$ we define the SVDMix operator by linearly blending the factors:  
\begin{equation}
  U_{\alpha}= \alpha U_{\text{young}} + (1-\alpha)U_{\text{old}},
  \quad\text{etc.}
  \label{eq:blend}
\end{equation}
and set
\begin{equation}
  B_{\text{fused}}(\alpha)=U_{\alpha}\Sigma_{\alpha}V_{\alpha}^{\!\top},
  \qquad
  A_{\text{fused}}(\alpha)=P_{\alpha}\Lambda_{\alpha}Q_{\alpha}^{\!\top}.
  \label{eq:bfaf}
\end{equation}
The final weight update used by the diffusion model is
\begin{equation}
  \Delta\theta_{\text{fused}}(\alpha)
  =B_{\text{fused}}(\alpha)\,A_{\text{fused}}(\alpha).
  \label{eq:dtheta_fused}
\end{equation}

We denote by $\|\cdot\|_{\mathrm F}$ the Frobenius norm and
recall two elementary facts:
(i) every matrix entry obtained from linear interpolation(Eq.~\ref{eq:blend}) is an \emph{affine} function of $\alpha$, hence continuous;
(ii) matrix addition and multiplication are linear and therefore continuous with respect to the
Frobenius norm.


The mapping
$
f:[0,1]\to\mathbb R^{m\times n},\;
\alpha\mapsto\Delta\theta_{\emph{fused}}(\alpha)
$
defined in~\eqref{eq:dtheta_fused}
is continuous on the closed interval $[0,1]$.

Fix any $\alpha_0\in[0,1]$ and let $\{\alpha_k\}\subset[0,1]$ be
a sequence with $\alpha_k\to\alpha_0$.
Because each entry of the blended factors
$U_{\alpha},\Sigma_{\alpha},V_{\alpha},P_{\alpha},\Lambda_{\alpha},Q_{\alpha}$
is an affine (hence continuous) function of $\alpha$,
we have
\begin{equation}
  U_{\alpha_k}\!\xrightarrow[k\to\infty]{} U_{\alpha_0},
  \quad
  \Sigma_{\alpha_k}\!\xrightarrow[k\to\infty]{} \Sigma_{\alpha_0},
  \quad
  \text{etc.}
  \label{eq:factor-conv}
\end{equation}
in the entry-wise (and thus Frobenius) sense.
By continuity of matrix multiplication,
\begin{equation}
\begin{aligned}
B_{\text{fused}}(\alpha_k)
    &= U_{\alpha_k}\Sigma_{\alpha_k}V_{\alpha_k}^{\!\top}
        \\ &\xrightarrow[k\to\infty]{}
       U_{\alpha_0}\Sigma_{\alpha_0}V_{\alpha_0}^{\!\top}
       \;=\; B_{\text{fused}}(\alpha_0).
\end{aligned}
\end{equation}
The same holds for $A_{\text{fused}}(\alpha_k)$.
Finally, using bilinearity of the matrix product,
\begin{equation}
\begin{aligned}
\Delta\theta_{\text{fused}}(\alpha_k)
    &= B_{\text{fused}}(\alpha_k)\,A_{\text{fused}}(\alpha_k)
        \\ &\xrightarrow[k\to\infty]{}
       B_{\text{fused}}(\alpha_0)\,A_{\text{fused}}(\alpha_0)
       \;=\; \Delta\theta_{\text{fused}}(\alpha_0).
\end{aligned}
\end{equation}
Hence $f$ is sequentially continuous, which on a
metric space implies continuity.


Because network weights vary continuously with~$\alpha$, the diffusion model’s output varies smoothly as the target age parameter is swept.

\section{Sensitivity Experiment}
\begin{figure*}[!htbp]
    \centering
    \includegraphics[width=0.9\textwidth]{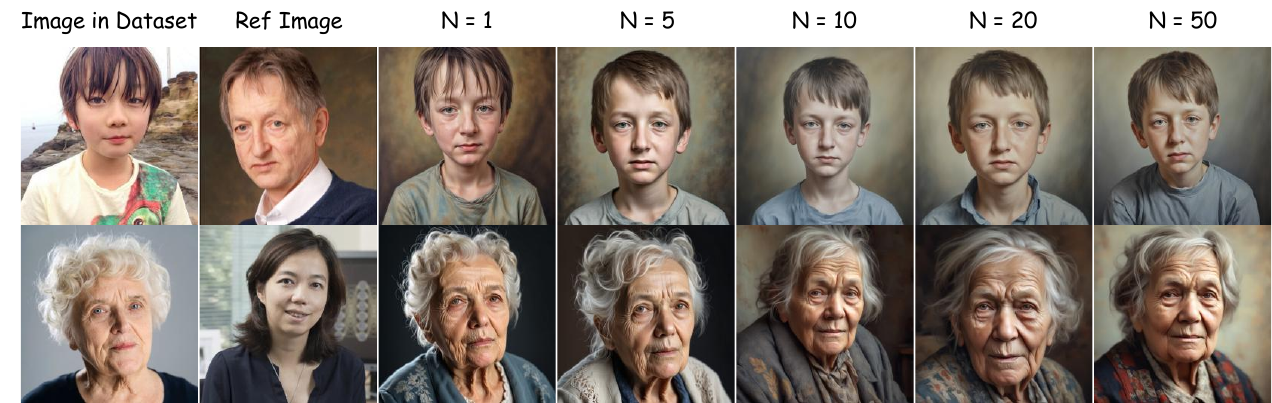}
    \caption{\textbf{Sensitivity experiment.} We vary training size $N \in \{1, 5, 10, 20, 50\}$ to assess data dependence. Identity drift is evident when $N \leq 5$, but stabilizes for $N \geq 10$.
    }
    \label{fig:sensitivity}
\end{figure*}

To investigate the appropriate amount of training data, we conducted a sensitivity analysis for both small ($N<10$) and larger values of $N$.
Specifically, we experimented with $N \in \{1, 5, 10, 20, 50\}$.
The smaller datasets were strict subsets of the larger ones to eliminate discrepancies due to data inconsistency.

\begin{table}[!htbp]
\centering
\caption{Sensitivity experiment metrics. MAE and age deviation at target ages 15 and 75 across different $N$.}
\label{tab:sensitivity}
\begin{tabular}{@{}ccccc@{}}
\toprule
 $N$ & 15 $\overline{\text{Age}}_{|\delta|\downarrow}$ & 15 MAE$\downarrow$ & 75 $\overline{\text{Age}}_{|\delta|\downarrow}$ & 75 MAE$\downarrow$ \\ \midrule
1  & $20.51_{5.51}$ & 6.73 & $70.35_{4.65}$ & 6.89 \\
5  & $11.95_{3.05}$ & 3.05 & $79.24_{4.24}$  & 4.40 \\
10 & $\underline{16.79}_{\underline{1.79}}$ & \textbf{2.82} & $\underline{78.82}_{\underline{3.82}}$ & \underline{3.98} \\
20 & $12.31_{2.69}$ & 2.84 & $\textbf{77.36}_{\textbf{2.36}}$  & \textbf{3.81} \\
50 & $\textbf{16.52}_{\textbf{1.52}}$  & 3.51 & ${79.01}_{{4.01}}$  & 4.09 \\ \bottomrule
\end{tabular}
\end{table}

The qualitative results are illustrated in Fig.~\ref{fig:sensitivity}. 
As shown, when the dataset is small ($N \leq 5$), the method is more susceptible to the influence of individual images, leading to identity drift—i.e., the generated identities tend to resemble the images in the training set.
In contrast, for $N \geq 10$, this issue is largely mitigated.
Quantitative comparisons are provided in Table~\ref{tab:sensitivity}, where it is evident that the model performs significantly better when $N \geq 10$. 
This indicates that with limited data ($N \leq 5$), the model's understanding of age is still insufficient.
Considering both the performance gain from additional data and the scarcity of high-quality age-labeled data, we find $N = 10$ to be a reasonable and practical choice.

\section{More Quality Experiments}

\begin{figure*}[!htbp]
    \centering
    \includegraphics[width=0.8\textwidth]{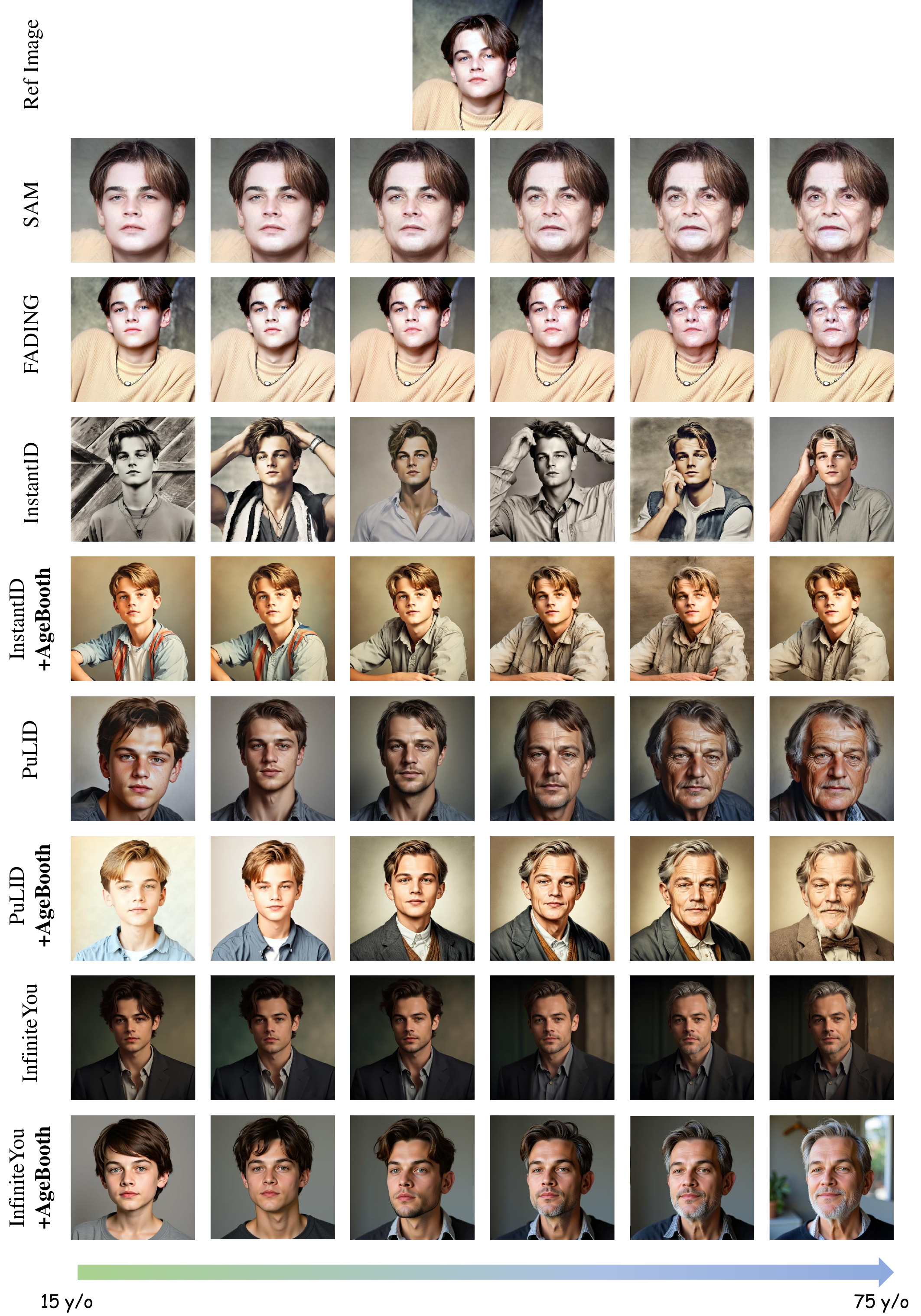}
    \caption{Quality comparison of interpolation results across all age groups}
    \label{fig:15_to_75}
\end{figure*}

\begin{figure*}[!htbp]
    \centering
    \includegraphics[width=0.8\textwidth]{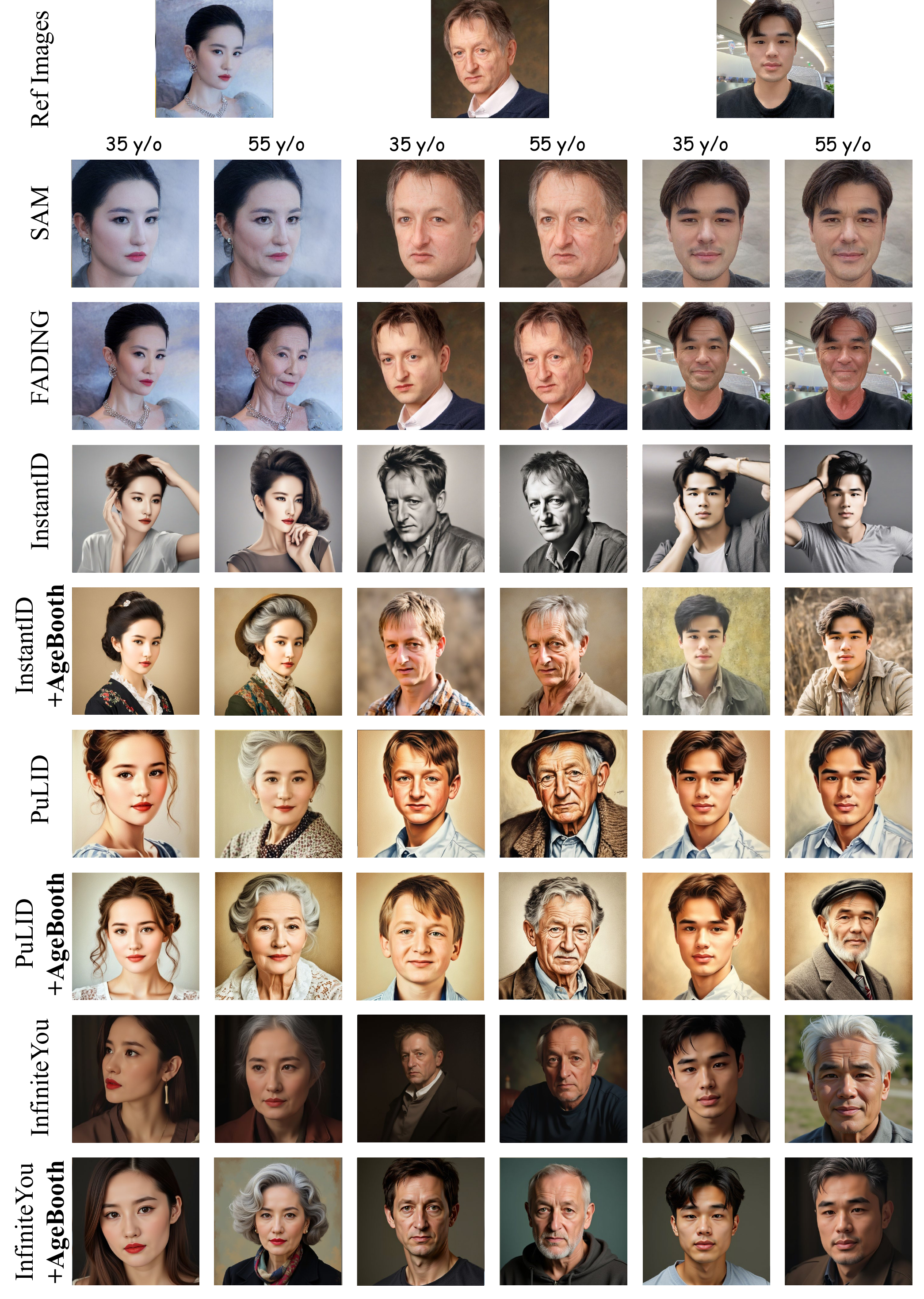}
    \caption{Quality comparisons of interpolation results at target ages 35 and 55.}
    \label{fig:more_q_35_55}
\end{figure*}

To further validate the effectiveness of our method beyond the anchor ages of 15 and 75, we conduct additional experiments at intermediate ages. 
These settings are essential for evaluating the interpolation ability of age transformation models. 
As shown in Fig.~\ref{fig:more_q_35_55},~\ref{fig:15_to_75} , we compare our AgeBooth method with several baselines at these age targets. 
We observe that integrating AgeBooth improves both visual quality and identity preservation. 
Compared to existing age-editing methods, AgeBooth generates faces with clearer details and better alignment with the target age. 
These findings further support the superiority of our approach in mid-age scenarios. 

\section{Extended Age‐Range Evaluation}\label{app:wideage}

\begin{figure*}[!htbp]
    \centering
    \includegraphics[width=0.9\textwidth]{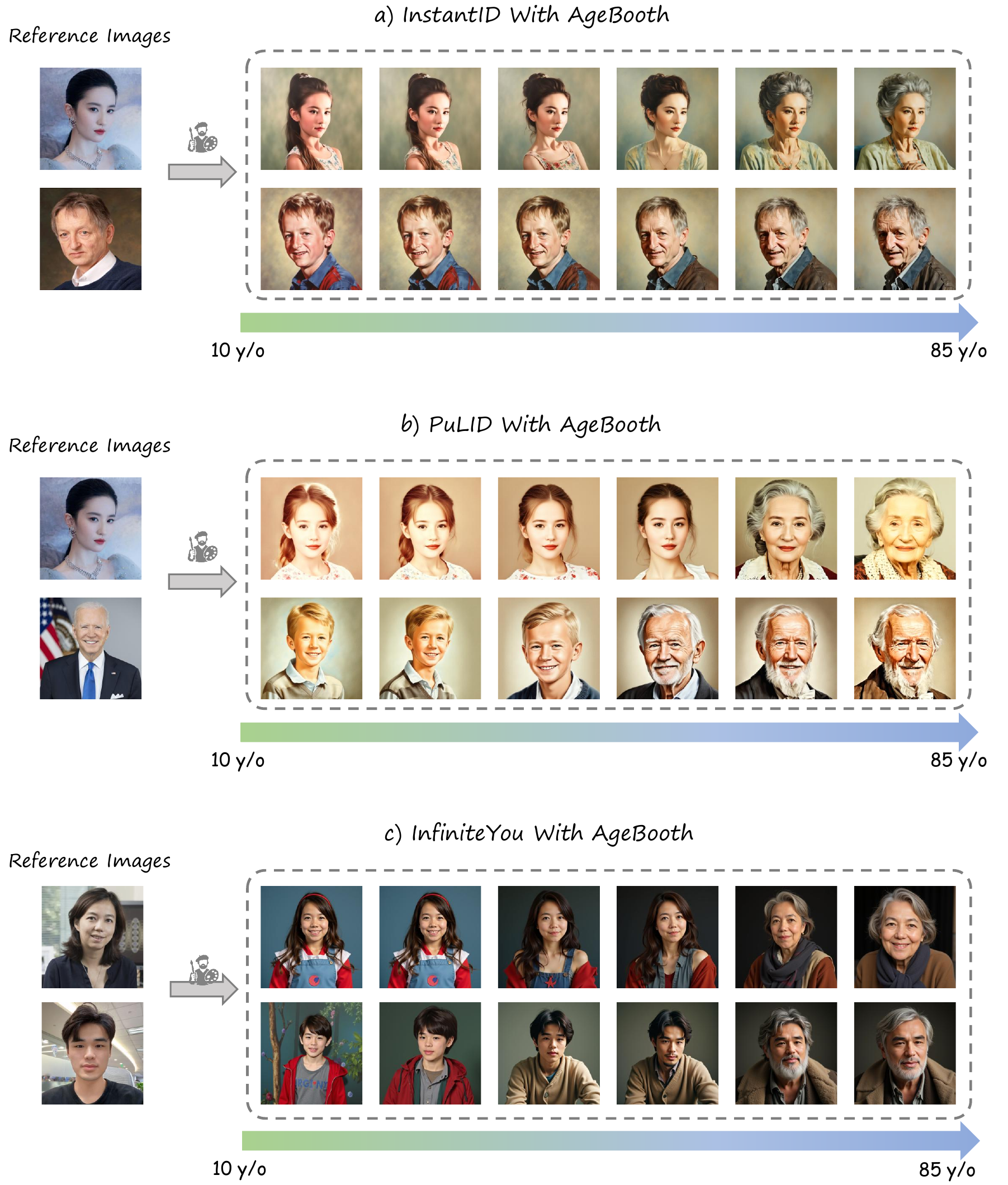}
    \caption{
    \textbf{AgeBooth over a broad age span (10–85\,yr).}  
    For each backbone we synthesise identity‑consistent portraits at six target ages between 10 and 85 years. (a) InstantID with AgeBooth, (b) PuLID with AgeBooth, and (c) InfiniteYou with AgeBooth.
    }

    \label{fig:10-85}
\end{figure*}

To verify that AgeBooth pipeline (Sec.~\ref{sec:age-finetuning} and  Sec.~\ref{subsec:fusions}) remains usable when the requested age falls outside the 15\,yr / 75\,yr anchor points, we further ran an additional evaluation covering the full adolescent‑to‑elderly span. No extra LoRA was trained and no hyper‑parameter was changed during inference.  We only vary the interpolation coefficient $\alpha$ (Eq.~8) and the blended prompt, exactly as in the main paper. The results are shown in Fig.~\ref{fig:10-85}.Visually, AgeBooth continues to produce smooth, realistic transitions in facial morphology, hair colour, and skin texture.
These observations confirm that a two‑anchor AgeBooth model can be directly deployed for much broader age requests, sparing users from training additional age‑specific adapters.

\section{More Visual Results}\label{app:visual_res}

\begin{figure*}[!htbp]
    \centering
    \includegraphics[width=0.9\textwidth]{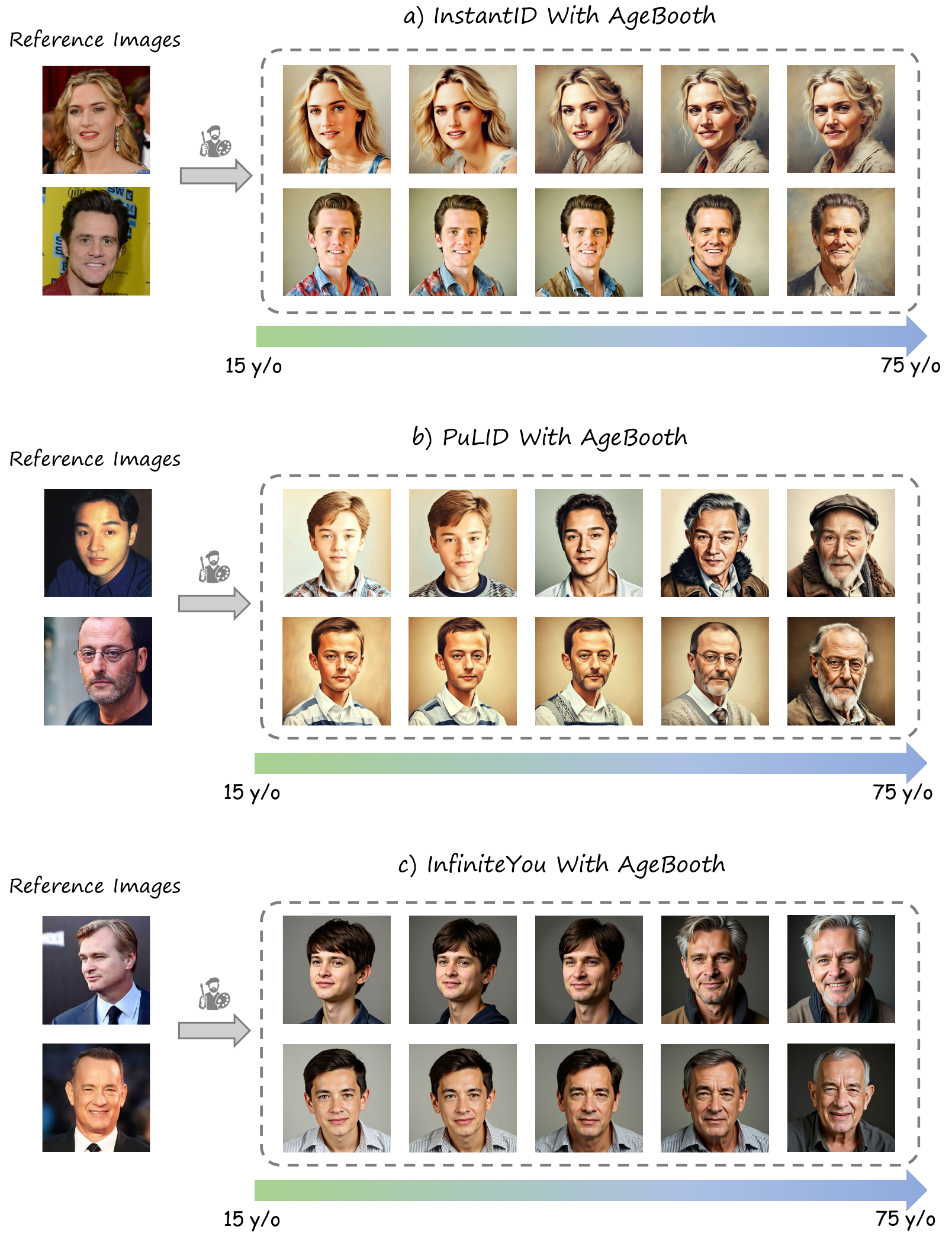}
    \caption{
    \textbf{Additional results generated by AgeBooth.} Portraits of a person across different age stages: (a) portraits generated by InstantID with AgeBooth; (b) portraits generated by PuLID with AgeBooth; (c) portraits generated by InfiniteYou with AgeBooth;
    }
    \label{fig:more_visuals}
\end{figure*}

\begin{figure*}
    \centering
    \includegraphics[width=\textwidth]{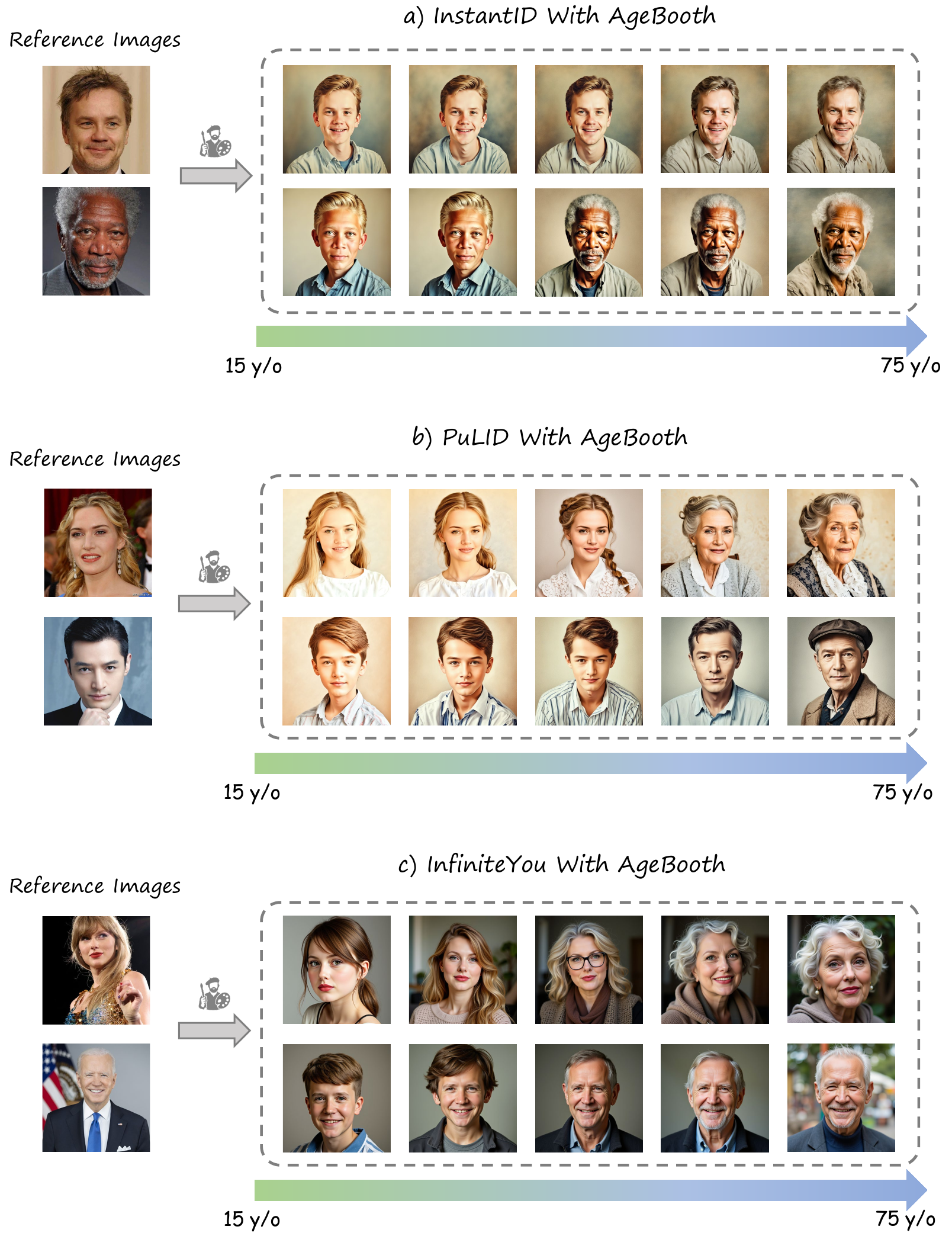}
    \caption{
    \textbf{Additional results generated by AgeBooth.} Portraits of a person across different age stages: (a) portraits generated by InstantID with AgeBooth; (b) portraits generated by PuLID with AgeBooth; (c) portraits generated by InfiniteYou with AgeBooth;
    }
    \label{fig:more_visuals1}
\end{figure*}

Following a similar setup as in Fig.~\ref{fig:teaser}, we present additional visualizations to further demonstrate our results.
Fig.~\ref{fig:more_visuals} and~\ref{fig:more_visuals1} show more results using AgeBooth on different base models(InstantID~\cite{wang2024instantid}, PuLID~\cite{guo2024pulid} and InfiniteYou~\cite{jiang2025infiniteyou}), illustrating age progression from 15 to 75 (left to right).
With our method, the models can generate multiple age stages with smooth transitions between them.

\section{Prompt Control}\label{app:prompt_control}
\begin{figure*}[!htbp]
    \centering
    \includegraphics[width=\textwidth]{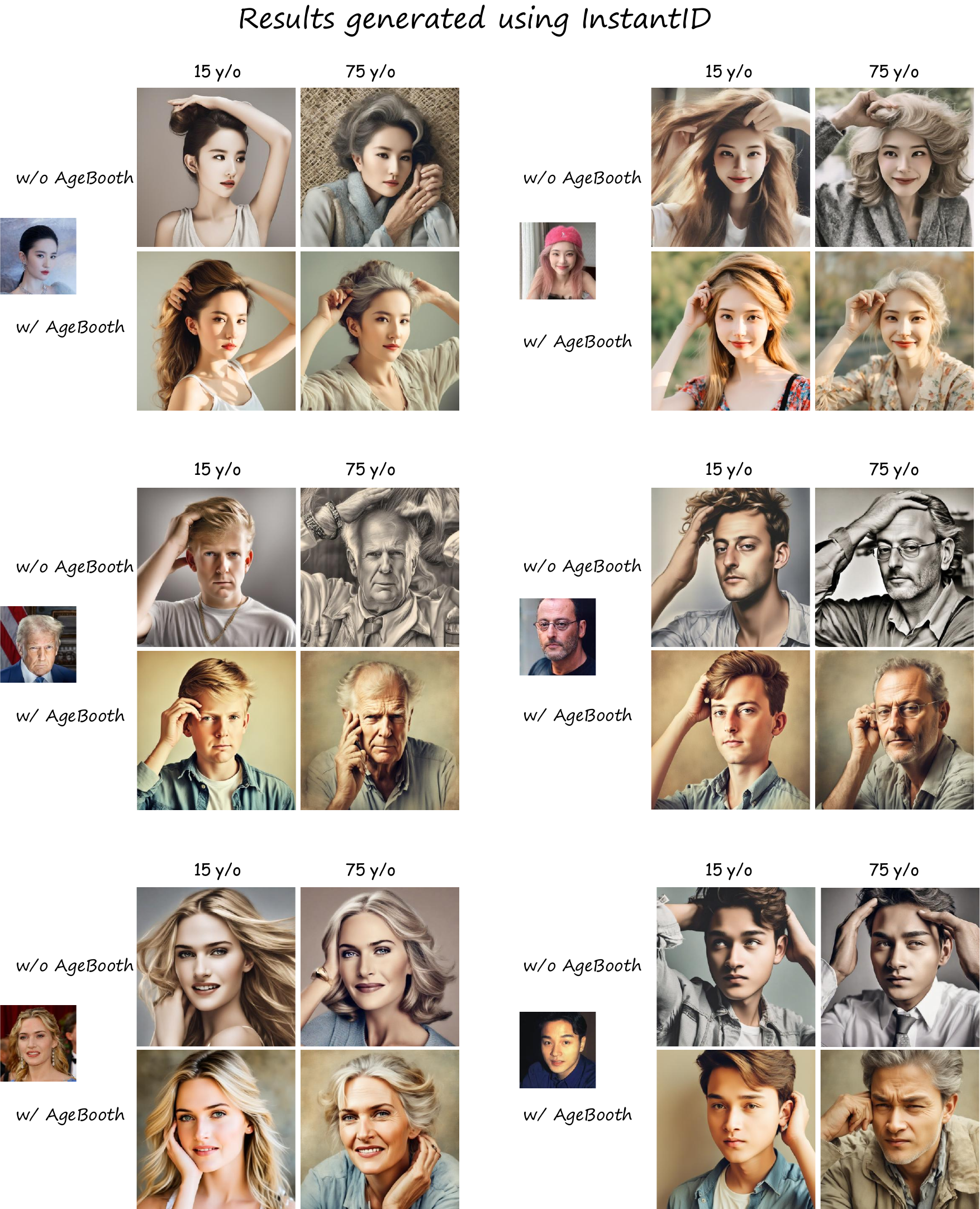}
    \caption{
    \textbf{Results generated using InstantID.} ``touching hair'' is used in prompt for generation.
    }
    \label{fig:prompt_control_instantid}
\end{figure*}

\begin{figure*}[!htbp]
    \centering
    \includegraphics[width=\textwidth]{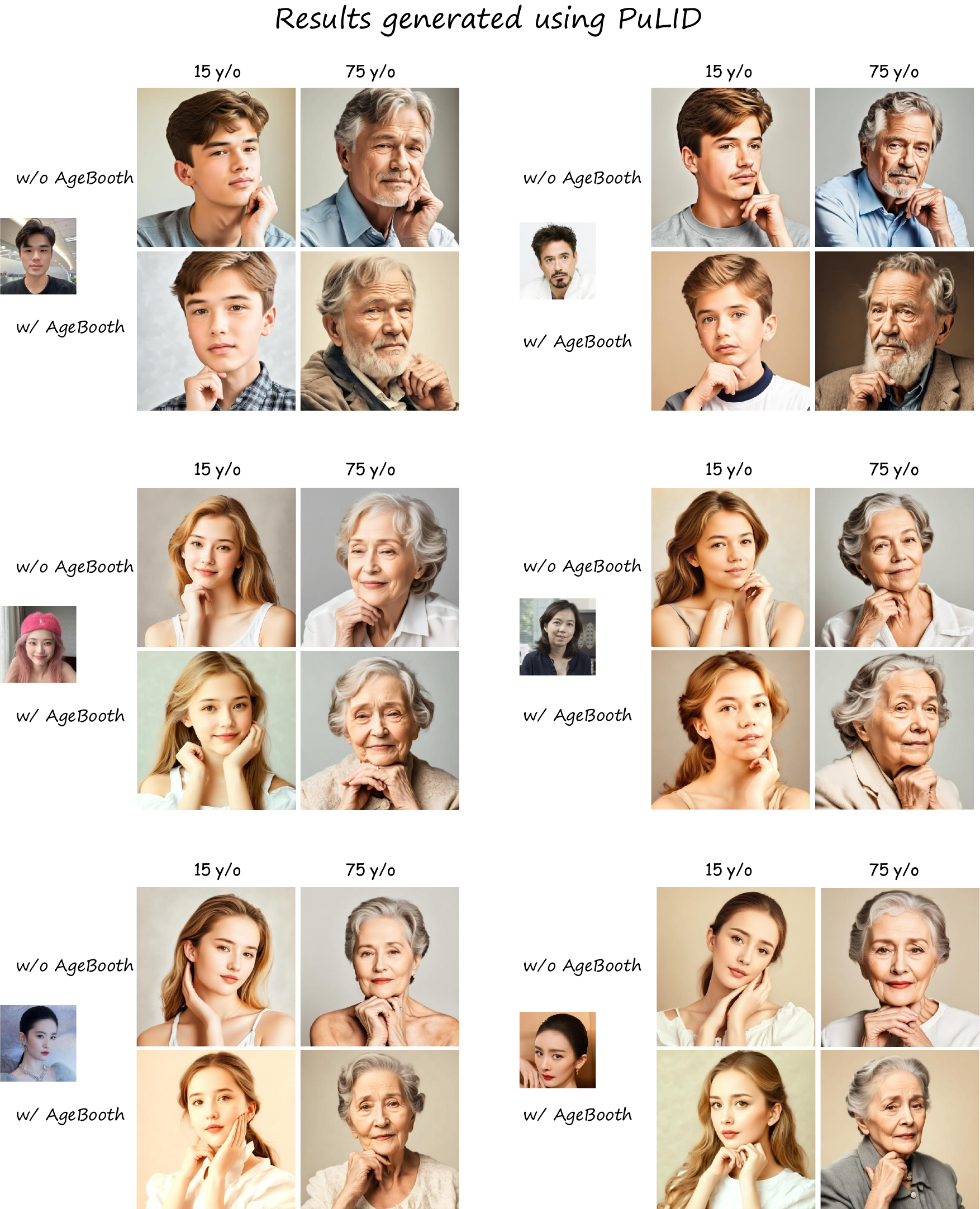}
    \caption{
    \textbf{Results generated using PuLID.} ``hand under chin'' is used in prompt for generation.
    }
    \label{fig:prompt_control_pulid}
\end{figure*}

\begin{figure*}[!htbp]
    \centering
    \includegraphics[width=\textwidth]{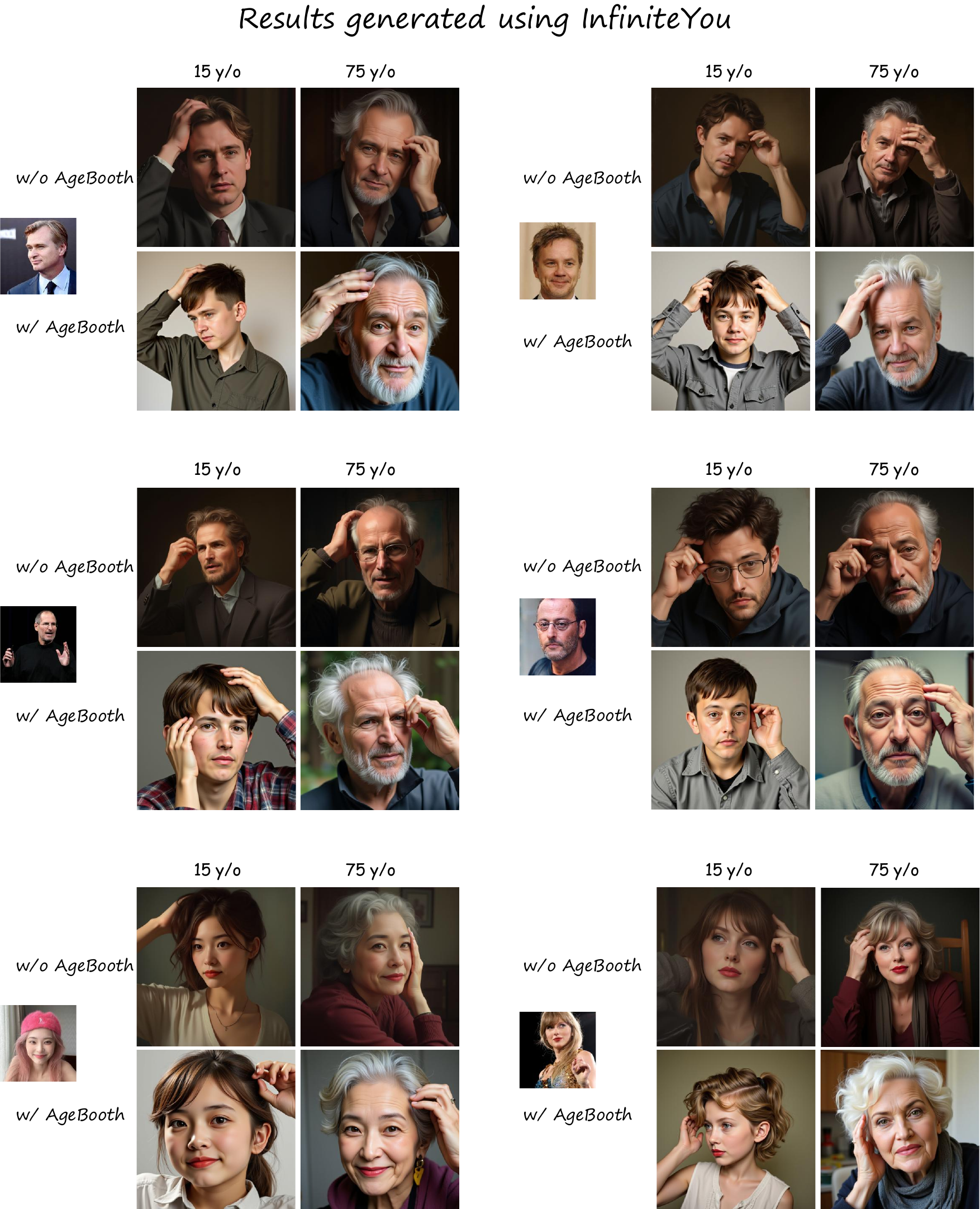}
    \caption{
    \textbf{Results generated using InfiniteYou.} ``scratching head'' is used in prompt for generation.
    }
    \label{fig:prompt_control_infinte}
\end{figure*}

We conduct a series of carefully designed experiments to thoroughly evaluate the controllability and flexibility of our method in generating images with diverse characteristics. 
Specifically, we aim to assess whether our approach can not only maintain consistent identity and age transformation, but also modify the subject’s actions,  and integrate novel elements as guided by prompt-based adjustments. 
As shown in Figs.~\ref{fig:prompt_control_instantid},~\ref{fig:prompt_control_pulid} and~\ref{fig:prompt_control_infinte}, after altering the input prompts, our method demonstrates a strong ability to preserve the intended age progression while simultaneously incorporating the specified changes in actions, and new elements.

\section{More Ablation on Interpolation Method}

\begin{figure*}[!htbp]
    \centering
    \includegraphics[width=0.9\textwidth]{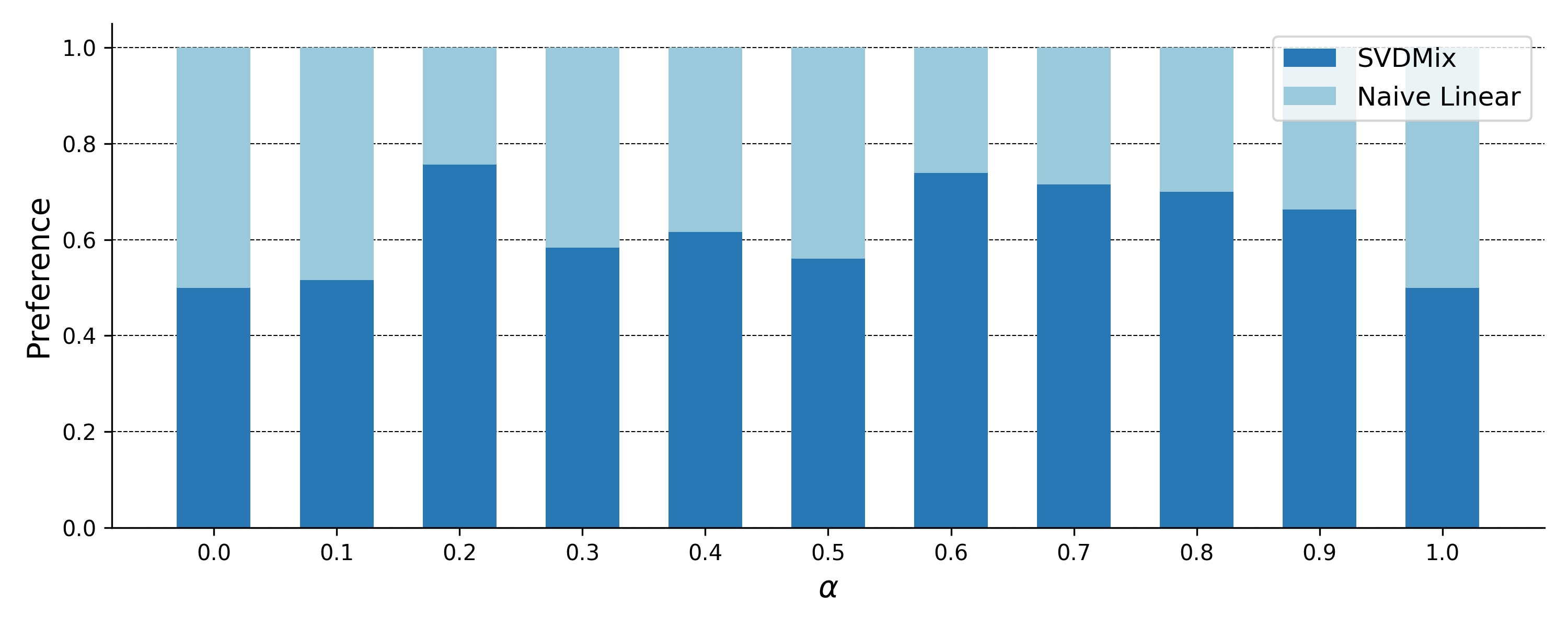}
    \caption{Preference comparison between SVDMix and naive linear interpolation across different interpolation weights $\alpha$.}
    \label{fig:ablation_preference}
\end{figure*}

To further validate the effectiveness of our interpolation strategy, we conduct a detailed comparison between \textbf{SVDMix} and the \textbf{naive linear interpolation} baseline.
We employ the Qwen2VL-7B~\cite{Qwen2VL,Qwen-VL} model to assess the quality of interpolated images along three critical dimensions: \textit{age accuracy}, \textit{consistency}, and \textit{image quality}. 
For age accuracy, we prompt the model with the interpolated image and a target reference age, and evaluate which method yields a result closer to the intended age.
For consistency, we query the model with an intermediate interpolated image and the original reference image, asking which method maintains a more faithful depiction of identity while achieving the target age.
For image quality, we similarly prompt the model to compare outputs from both methods. The final preference score at each interpolation weight $\alpha$ is obtained by aggregating the evaluations across these three aspects.
As shown in Figure~\ref{fig:ablation_preference}, the SVDMix-based method consistently outperforms naive linear interpolation, demonstrating superior control over age modulation, better identity preservation, and improved visual fidelity.

\section{More Details on the Evaluation Settings}\label{app:settings}
We random select images from CelebHQ~\cite{karras2018progressive} and cross-age image pairs from AgeDB~\cite{moschoglou2017agedb} for age and identity comparisons.
For age comparison, the CelebHQ images are used as references.
We select reference images that exhibit a significant age gap from the target.
Middle-aged images are used as references for the 15 and 75 age groups, while young and old images are used for the 45 age group.
Then we measure the gap between generated and target ages.
For identity comparison, AgeDB pairs are used, with 45-year-old images as references and 15 or 75-year-old images as comparisons.
Models generate target-age faces from the reference, and identity similarity is computed against the comparison images.

\end{document}